\documentclass[nonacm,sigconf]{acmart}
\AtBeginDocument{%
  }

\usepackage[table,xcdraw]{xcolor}
\usepackage{tabularray}
\usepackage{booktabs}
\usepackage{multirow}
\usepackage{array}
\usepackage{makecell}

\usepackage{hhline}
\usepackage{pifont}
\newcommand{\cmark}{\ding{51}}
\newcommand{\xmark}{\ding{55}}

\begin{document}

\title{Capture Timing-Attention of Events in Clinical Time Series}

\author{Jia Li}
\email{jiaxx213@umn.edu}
\affiliation{%
  \institution{Department of Surgery; Department of Computer Science, U of M}
  \city{Minneapolis}
  \state{MN}
  \country{USA}
}

\author{Yu Hou}
\email{hou00127@umn.edu}
\affiliation{%
  \institution{Department of Surgery, U of M}
  \city{Minneapolis}
  \state{MN}
  \country{USA}
}

\author{Rui Zhang}
\email{ruizhang@umn.edu}
\affiliation{%
  \institution{Department of Surgery, U of M}
  \city{Minneapolis}
  \state{MN}
  \country{USA}
}
\begin{abstract}

The contemporary paradigm of trajectory learning operates fundamentally at the level of group dynamics, systematically reducing individual-level complexity to fit group-level models, thus rendering effective patient subtyping difficult and individual-level modeling largely out of reach. We propose a data-driven paradigm that introduces a dedicated individual-level temporal variable to capture \emph{Timing Attention} (i.e., the degree of concentration of an event's timing distribution across the patient cohort), thereby rendering timing a \emph{computable dimension} that enables individualized temporal features in trajectory learning.

Instantiated as the Level-of-Individual Time Transformation (LITT) and applied to longitudinal EHR data from 3,276 breast cancer patients, the proposed paradigm demonstrates, for the first time to our knowledge: (1) automatic discovery of clinically significant patient trajectories, and (2) counterfactual timing deduction, that is, a \emph{What-If Machine}. Both results are purely data-driven, requiring no prior domain knowledge. LITT further achieves strong performance on timing prediction and survival analysis tasks.
\end{abstract}



\keywords{Timing Attention, Trajectory Discovery, What-If Machine, Precise Medicine}


\maketitle

\section{Introduction}

\begin{figure*}[h]
  \includegraphics[width=0.95\textwidth]{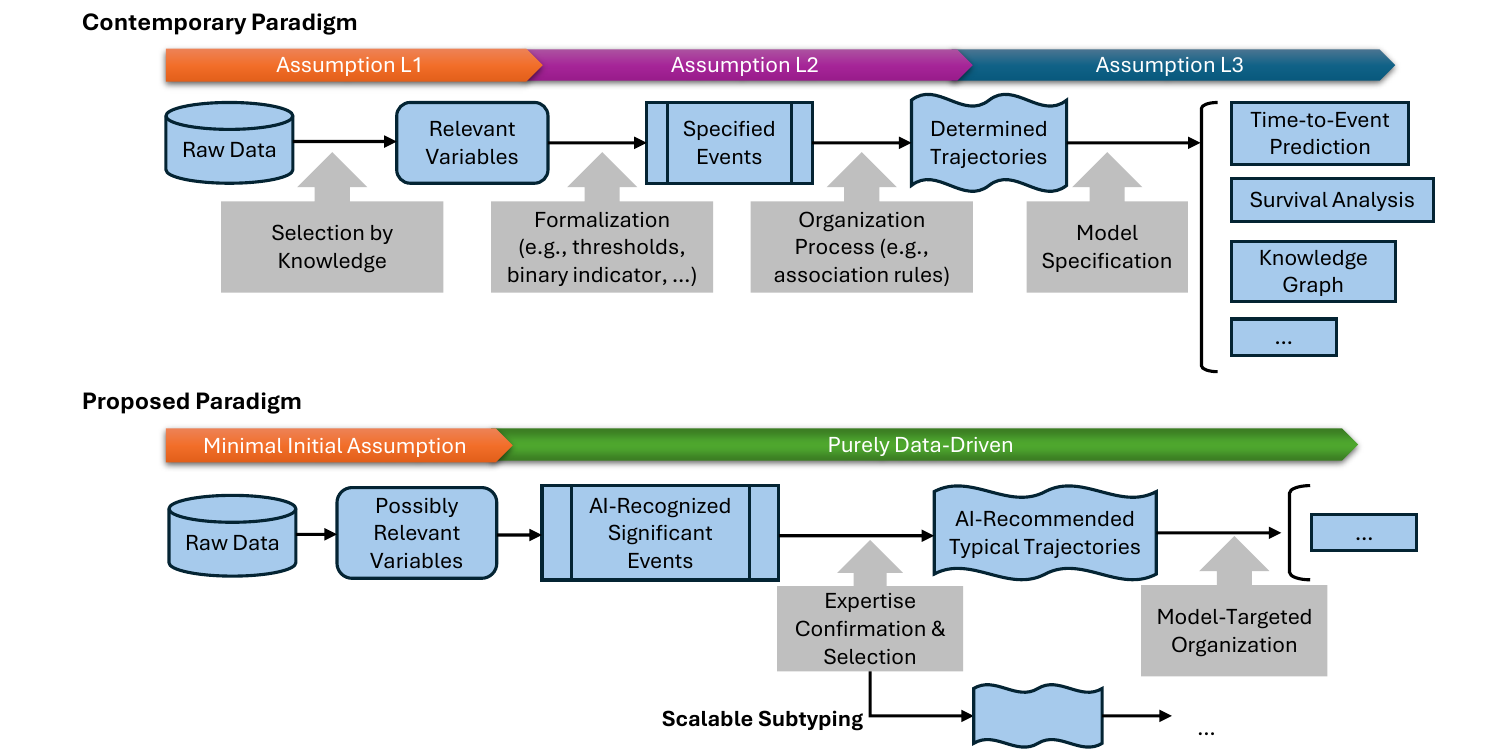}
  \vspace{-1mm}
    \caption{Comparison of the current and proposed paradigms for AI-based temporal trajectory modeling. The current paradigm implicitly relies on three layered assumptions: (L1) relevant variables are selected based on prior domain knowledge and assumed to fully represent the raw data; (L2) events are defined by discretizing continuous values based on domain experience; (L3) model specification encodes additional assumptions according to the targeted task. The proposed paradigm, by contrast, operates in a purely data-driven manner, whose feasibility is demonstrated by the method introduced in this work.}
\vspace{-1mm}
  \label{fig:paradigm}
\end{figure*}

Electronic health records (EHRs) store patient clinical events longitudinally, thereby forming unique patient trajectories. These trajectories encode rich patient-specific information, including diagnosis history, laboratory values, drug administration, and treatment plans, that is highly predictive of disease progression and clinical outcomes \cite{jensen2014temporal, xiao2018opportunities}. Effectively modeling such trajectories to uncover personalized temporal information is therefore critical for advancing precision medicine \cite{sitapati2017integrated, sanchez2022causal}.

The contemporary paradigm of trajectory learning, however, operates fundamentally at the level of group dynamics. Although recently developed AI solutions can leverage far richer longitudinal data than traditional methods \cite{allam2021analyzing}, they still lack the individualized representational capacity necessary to yield patient-specific interpretations. 
More critically, as illustrated in Figure \ref{fig:paradigm}, the assumptions underlying this paradigm systematically reduce individual-level complexity in order to fit group-level models — at the cost of potentially discoverable patient-specific knowledge. As a result, such models can hardly support effective patient subtyping, let alone modeling at the level of individual \cite{landi2020deep, eskofier2023predictive}.

We propose a data-driven paradigm for AI-based temporal trajectory modeling (Figure \ref{fig:paradigm}), in which AI first discovers significant events and trajectories directly from the source data; domain expertise is subsequently applied to refine and select among these discoveries according to the modeling purpose — enabling scalable patient subtyping down to the individual level.

This naturally calls for a data-driven approach to evaluating the significance of candidate events across the patient cohort. We introduce the \emph{relative timing} of patients as a learnable distribution over an event's possible timing, whose degree of concentration directly characterizes the \emph{\textbf{Timing Attention}} of that event — a high concentration indicating that the event represents a decisive, shared moment across the cohort, and a low concentration indicating temporal dispersion and lower collective significance.

The patient's relative timing is represented by a dedicated individual-level variable, learned through backpropagation jointly with, but independently parameterized from, the shared group-level model components. 
This formulation renders \emph{timing a \textbf{computable dimension}}, enabling individualized temporal representation at the patient level. We term this approach the Level-of-Individual Time Transformation (LITT).

Applied to longitudinal EHR data from 3,276 breast cancer patients, LITT empirically demonstrates: (1) automatic discovery of clinically significant patient trajectories, and (2) counterfactual timing deduction — estimating when a cardiac event would have occurred had a positive patient remained negative. Both results are purely data-driven, requiring no prior domain knowledge, and represent, to our knowledge, the first such demonstrations in the machine learning literature. Additionally, LITT achieves strong performance on timing prediction and survival analysis tasks, as demonstrated in the experimental evaluation.
\subsection{Related Works}

Based on their temporal representations, we categorize trajectory learning methods as shown in Table \ref{tab:category}. 

In the traditional framework, timing is not treated as a continuous variable $t$ but rather as an input sequence of absolute timestamps $\{...,t_i, t_{i+1},...\}$.
As a typical example, time-to-event prediction focuses on specific risk scores associated with a predetermined timestamp of interest \cite{allam2021analyzing, esteva2019guide}, wherein the training process involves no explicit temporal variables. 
Similarly, classical survival analysis treats the time segments delimited by timestamps as discrete modeling objectives \cite{wang2019machine, wang2022survtrace, lee2018deephit, katzman2018deepsurv}.
Although AI-based methods have substantially enhanced modeling effectiveness across these tasks \cite{xiao2018opportunities, choi2016doctor, lipton2015learning, wiegrebe2024deep}, they do not alter the fundamental nature of the framework: absolute timing is discretized in advance, and temporal-relative modeling is eliminated by design.

Beyond prediction tasks, some dynamic modeling approaches have been applied to clinical trajectory data \cite{schulam2015clustering, hirano2002mining, bossa2023multidimensional}, reframing the temporal trajectories of specific variables with established clinical interpretations as a spatial problem, so that time serves merely as a coordinate axis rather than as a quantity to be modeled. 
A further reduction is employed in knowledge graph-based methods, which collapse timing entirely into event ordering — learning structured associations among ontological terms without retaining any temporal magnitude \cite{jensen2014temporal, beck2016diagnosis}.

As RNNs have been increasingly adopted for longitudinal EHR modeling, the irregularly sampled nature of real-world clinical data has emerged as a central challenge \cite{turkson2021handling}, since standard RNNs presuppose uniform time steps, thereby motivating a growing body of work on temporally adaptive architectures. Two broad strategies have emerged:
The first, known as \emph{time-aware RNNs}, adjusts the hidden-state update mechanism within the recurrent cell to account for irregular time gaps between patient visits. Representative methods include T-LSTM \cite{baytas2017patient}, GRU-D \cite{che2018recurrent} and TA-RNN \cite{al2024ta}.
The second strategy, \emph{time representation learning}, encodes time as an explicit input embedding without modifying the recurrent cell. Time2Vec \cite{kazemi2019time2vec} maps timestamps to sinusoidal and linear components, while mTAN \cite{shukla2021multi} learns a continuous-time attention kernel to measure temporal similarity.

A limitation that has received insufficient attention is that such state-based adjustment typically imposes a Markov assumption on the trajectory, a constraint well-suited to physical dynamic systems but rarely warranted in clinical data.
In practice, combined treatments may occur in varying orders or absences across patients, yet converge toward equivalent clinical outcomes.
From this perspective, the \emph{ODE-based scheme} offers a more principled alternative \cite{rubanova2019latent, kidger2020neural}. They model trajectories as continuous, history-aware dynamics that are not constrained to the immediately preceding state.
However, this non-Markovian capability remains confined to the group level — shared continuous dynamics cannot, by construction, capture the temporal heterogeneity that distinguishes individual patients.
%

\begin{table*}[t]
\small
\centering
\begin{tabular}{lccllcllc}
 & \multicolumn{8}{l}{Marker *A = Non-Markovian} \\
 & \multicolumn{8}{l}{Marker *B = Timing Distribution \& Prediction} \\ \cline{2-9} 
\multicolumn{1}{l|}{} & \multicolumn{1}{c|}{\cellcolor[HTML]{D1D1D1}\textbf{Category}} & \multicolumn{3}{c|}{\cellcolor[HTML]{FFFFFF}Absolute Timing} & \multicolumn{3}{c|}{\cellcolor[HTML]{FFFFFF}\begin{tabular}[c]{@{}c@{}}Deterministic \\ Relative Timing\end{tabular}} & \multicolumn{1}{c|}{\cellcolor[HTML]{FFFFFF}\begin{tabular}[c]{@{}c@{}}Computable \\ Relative Timing\end{tabular}} \\ \cline{2-9} 
\multicolumn{1}{l|}{} & \multicolumn{1}{c|}{\cellcolor[HTML]{D1D1D1}\textbf{Perspective}} & \multicolumn{3}{c|}{Time-First} & \multicolumn{3}{c|}{State-First} & \multicolumn{1}{c|}{Event-First} \\ \cline{2-9} 
\multicolumn{1}{l|}{} & \multicolumn{1}{c|}{\cellcolor[HTML]{D1D1D1}\textbf{Solution}} & \multicolumn{1}{l|}{Discretize} & \multicolumn{1}{l|}{Spatialize} & \multicolumn{1}{l|}{\begin{tabular}[c]{@{}l@{}}Reduce to \\ Ordering\end{tabular}} & \multicolumn{1}{l|}{\begin{tabular}[c]{@{}l@{}}Time-Aware \\ RNN\end{tabular}} & \multicolumn{1}{l|}{\begin{tabular}[c]{@{}l@{}}Time \\ Representation\end{tabular}} & \multicolumn{1}{l|}{ODE-based} & \multicolumn{1}{l|}{\begin{tabular}[c]{@{}l@{}}Introduce  \\ Individual Level\end{tabular}} \\ \cline{2-9} 
\multicolumn{1}{l|}{} & \multicolumn{1}{c|}{\cellcolor[HTML]{D1D1D1}\textbf{Models}} & \multicolumn{1}{l|}{\textit{\begin{tabular}[c]{@{}l@{}}Time-to-Event Prediction,\\ Survival Analysis\end{tabular}}} & \multicolumn{1}{l|}{\textit{\begin{tabular}[c]{@{}l@{}}Specific \\ Dynamic\end{tabular}}} & \multicolumn{1}{l|}{\textit{\begin{tabular}[c]{@{}l@{}}Association\\ Graph\end{tabular}}} & \multicolumn{1}{l|}{\textit{\begin{tabular}[c]{@{}l@{}}T-LSTM, \\ GRU-D, TA-RNN\end{tabular}}} & \multicolumn{1}{l|}{\textit{\begin{tabular}[c]{@{}l@{}}Time2Vec, \\ mTAN\end{tabular}}} & \multicolumn{1}{l|}{\textit{\begin{tabular}[c]{@{}l@{}}Latent ODE, \\ Neural ODE\end{tabular}}} & \multicolumn{1}{l|}{\textit{\begin{tabular}[c]{@{}l@{}}LITT (this work)\end{tabular}}} \\ \cline{2-9} 
\multicolumn{1}{l|}{} & \multicolumn{1}{c|}{\cellcolor[HTML]{FFFFFF}\textbf{*A}} & \multicolumn{3}{c|}{\cellcolor[HTML]{FFCCC9}\xmark} & \multicolumn{2}{c|}{\cellcolor[HTML]{FFCCC9}\xmark} & \multicolumn{1}{c|}{\cellcolor[HTML]{BCFABB}\cmark} & \multicolumn{1}{c|}{\cellcolor[HTML]{BCFABB}\cmark} \\ \cline{2-9} 
\multicolumn{1}{l|}{} & \multicolumn{1}{c|}{\cellcolor[HTML]{FFFFFF}\textbf{*B}} & \multicolumn{3}{c|}{\cellcolor[HTML]{FFCCC9}\xmark} & \multicolumn{2}{c|}{\cellcolor[HTML]{FFCCC9}\xmark} & \multicolumn{1}{c|}{\cellcolor[HTML]{FFCCC9}\xmark} & \multicolumn{1}{c|}{\cellcolor[HTML]{BCFABB}\cmark} \\ \cline{2-9} 
\end{tabular}
\vspace{2mm}
\caption{Categorization of trajectory learning methods by temporal representation. Traditional methods operate on fixed absolute timestamps; existing deep learning models produce deterministic relative timing representations; LITT is the first to treat event timing as a learnable distribution rather than a fixed value.}
\label{tab:category}
\end{table*}

\begin{figure*}[h]
  \includegraphics[width=0.98\textwidth]{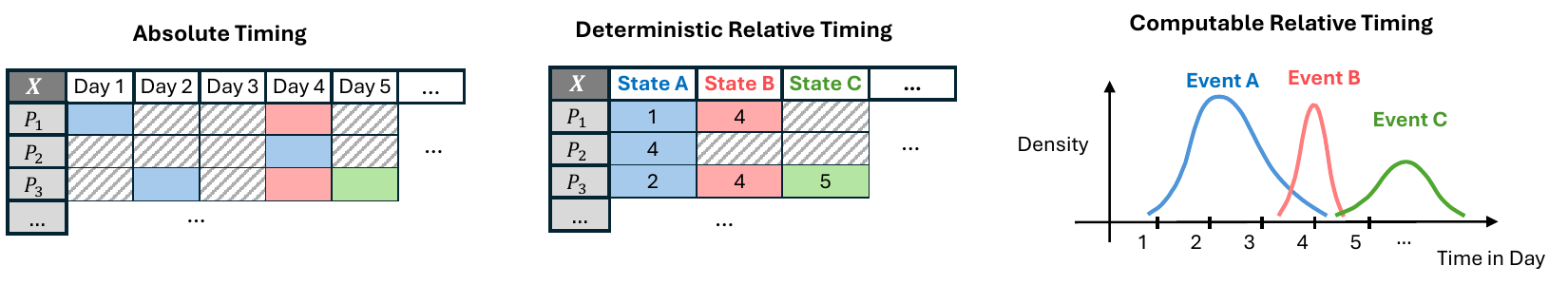}
  \vspace{-3mm}
  \caption{Comparison of \emph{absolute} timing, \emph{deterministic} relative timing, and \emph{computable} relative timing representations across patient trajectories, where $P_i$ $ (i=1,2,3,...)$ denotes individual patients.}
\vspace{-1mm}
  \label{fig:arrange}
\end{figure*}

\subsection{Computable Relative Timing}
In existing works, relative timing has been incorporated as a \emph{learnable} variable to better capture group-level dynamics, converging to a deterministic value upon training completion. 
Individualized temporal modeling, however, requires relative timing to be a \emph{computable} dimension to accommodate a distribution over possible timing values across patients rather than collapsing to a single determined value. 
Unlike \emph{timing-learnable}, \emph{timing-computable} denotes the strictly broader capability that additionally encompasses \emph{timing-predictable} — the latter being precisely the capacity that existing methods lack.

The absence of timing prediction in the literature is itself revealing: since $t$ appears as an input variable in the objective function, repositioning it as the outcome variable would seem, at first glance, sufficient to achieve timing prediction, that is, regression on event timing. 
The barrier is not technical sophistication but a missing foundational concept: a formal definition of the \emph{possible timing distribution} of an event, which must be established prior to any methodological discussion and upon which meaningful predictive evaluation can be constructed.

Figure \ref{fig:arrange} illustrates the three temporal representations corresponding to the categories in Table \ref{tab:category}. 
Absolute timing arranges time-series data at fixed time steps, resulting in substantial sparsity — a \emph{time-first} perspective in which the temporal grid precedes and constrains the data. 
Deterministic relative timing presupposes sequential common states that adequately represent shared dynamics across individuals — a \emph{state-first} perspective in which the state structure is assumed before the data is observed. 
Computable relative timing, by contrast, discovers events post-hoc from collective timing distributions rather than relying on predetermined event specifications — an \emph{event-first} perspective in which the data itself determines what constitutes a meaningful event. This deeper data-driven capability, while not fully realized in the present work, represents a promising direction for future research.

This work makes three major contributions.
\begin{itemize}
\item propose a data-driven paradigm for individual-level temporal trajectory modeling, realized as LITT (Level-of-Individual Time Transformation), empirically confirming the feasibility of timing computation without recourse to prior domain knowledge;
\item address a foundational conceptual gap by formally defining the \emph{distribution of possible timing} and introducing \emph{Timing Attention} as a principled measure of an event's temporal significance;
\item analyze the fundamental difference between LSTM and GRU with respect to timing transformation, demonstrating that LSTM's dedicated cell state is essential for timing computation whereas GRU's gating mechanism lacks this capability by construction.
\end{itemize}

\section{Level-of-Individual Time Transformation }

\subsection{Relative Time Transformation}

\begin{figure}
  \includegraphics[width=0.48\textwidth]{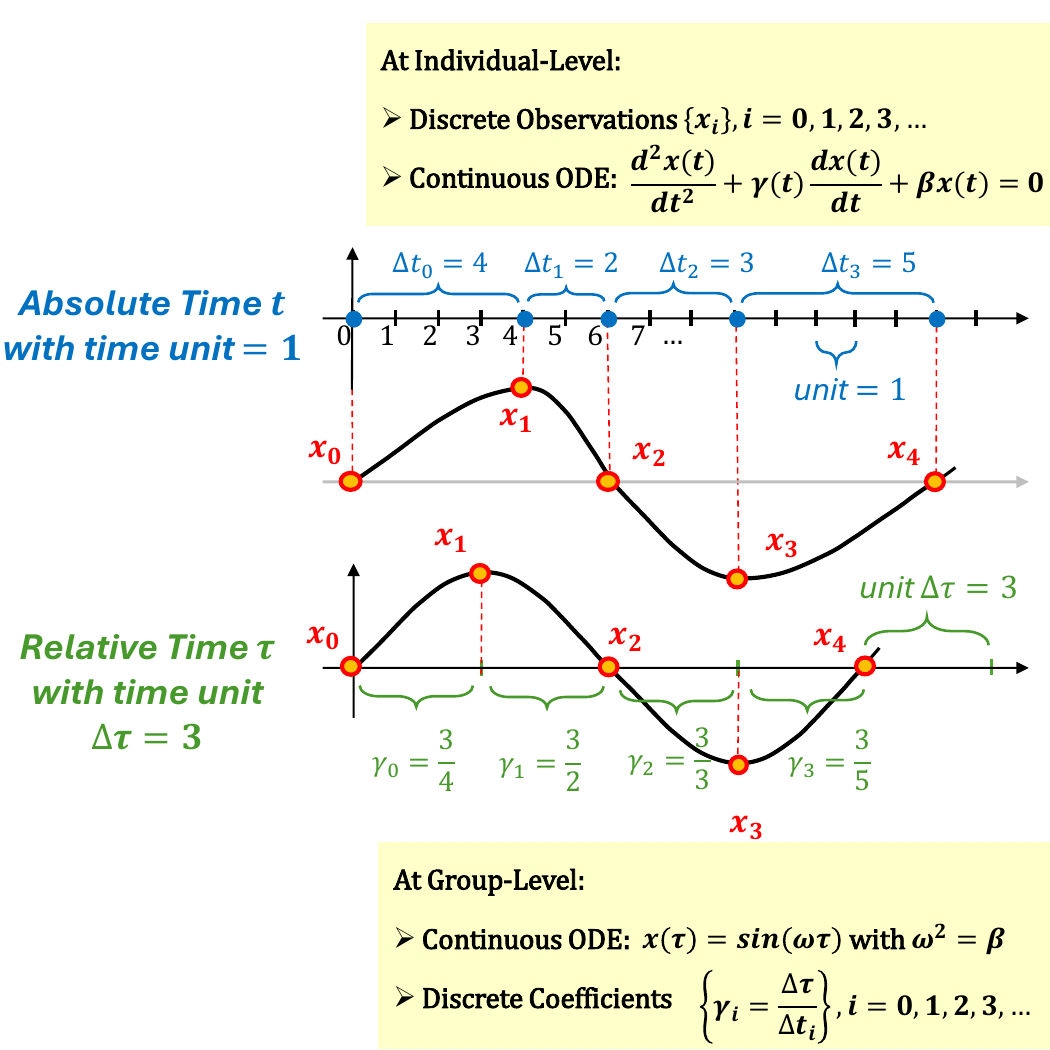}
  \vspace{-1mm}
  \caption{An illustration of the time transformation from absolute timeline $t$ to relative timeline $\tau$. Upon training completion, the dynamics on $\tau$ converge to a deterministic mapping encoding the shared group-level trajectory pattern, while the individual coefficient sequence $\{\gamma_i\}_{i=0}^{T}$ constitutes each patient's temporal feature — a single realization drawn from the relative timing distribution.}
\vspace{-1mm}
  \label{fig:gamma}
\end{figure}

Figure \ref{fig:gamma} illustrates step-wise time transformation using a simple sinusoidal curve as an individual's dynamic trajectory. The absolute time interval $  \Delta t  $ is transformed by the model into a relative interval $  \Delta \tau  $, and the relative timeline $  \tau  $ is gradually constructed by accumulating these sequential $  \Delta \tau  $ values.
Here, the derived coefficients $  \{\gamma_i\}  $ represent the sequential time-scaling rates governing this transformation.
Each patient possesses an individual coefficient sequence $  \{\gamma_i\}_{i=0}^{T}  $. Consequently, the model operates on an $  N \times T  $ coefficient matrix $  V  $, where $  \gamma_{ij} \in V  $ denotes the $  i  $-th time step's scaling coefficient for patient $  j  $ ($  j = 0, \dots, N  $).

The matrix $  V  $ is obtained by simultaneously optimizing across all $  N  $ individuals in the group, thereby enforcing a globally effective relative timeline $  \tau  $ that captures their shared group-level trajectory pattern.
Conversely, the original absolute timestamps encode individual-level trajectories, which can be recovered from the common dynamic on $  \tau  $ by applying the corresponding coefficients $  \{\gamma_{ij}\}_{i=0}^{T}  $ for each patient $  j  $.

Let $  x(\tau)  $ denote the continuous dynamic trajectory transformed from $  x(t)  $, which manifests as the observed sequence $  \{x_{ij}\}_{i=0}^{T}  $ for patient $  j  $.
The time-scaling function $  \gamma(t)  $ can then be expressed through a second-order ordinary differential equation (ODE):
\begin{align}
x''(t) + \gamma(t) x'(t) + \beta x(t) = 0, \quad \beta > 0 \label{eq2}
\end{align}
Our goal is to eliminate the first-order damping term $  \gamma(t) x'(t)  $, reducing the ODE to the undamped harmonic oscillator form $x''(t) + \beta x(t) = 0.$
Let $  \tau = \tau(t)  $ denote the desired time transformation, so that $  x(t) = x(\tau(t))  $. Applying the chain rule yields:
\begin{align*}
\textit{first order: } x'(t) &= \frac{dx}{dt} = \frac{dx}{d\tau} \frac{d\tau}{dt} = x'(\tau) \tau'(t) \\
\textit{second order: } x''(t) &= \frac{d}{dt} \left( x'(\tau) \tau'(t) \right) = x''(\tau) (\tau'(t))^2 + x'(\tau) \tau''(t)
\end{align*}

Substituting into Equation~(\ref{eq2}) gives:
\begin{align*}
x''(\tau) (\tau'(t))^2 + x'(\tau) \left[ \tau''(t) + \gamma(t) \tau'(t) \right] + \beta x(\tau) = 0 \\
\end{align*}

To eliminate the first-order damping term, we set the coefficient of $  x'(\tau)  $ to zero to obtain the solution of this time transformation:
\begin{align}
    \textit{In continuous: } \tau'(t) &= C e^{-\int \gamma(t) \, dt} \text{ with }  C \in \mathbb{R} \nonumber \\
    \textit{In discrete: } \frac{d\tau}{dt} &= C e^{-\sum^t_{i=0} \gamma_i} \text{ with }  C \in \mathbb{R} \label{eq3}
\end{align}

From Equation~(\ref{eq3}), achieving the desired transformation at step $  i  $ requires scaling all preceding absolute timestamps $  t_0, \dots, t_{i-1}  $ by the cumulative factors $  e^{-\gamma_0}, \dots, e^{-\gamma_{i-1}}  $, respectively.
Within a recurrent architecture, this is realized by a dedicated gate that computes $  e^{\gamma}  $ per step using absolute timestamps as input. The gate uses a specialized individual-level parameter that updates independently of the standard group-level recurrent modeling components. 

\begin{figure}
  \includegraphics[width=0.49\textwidth]{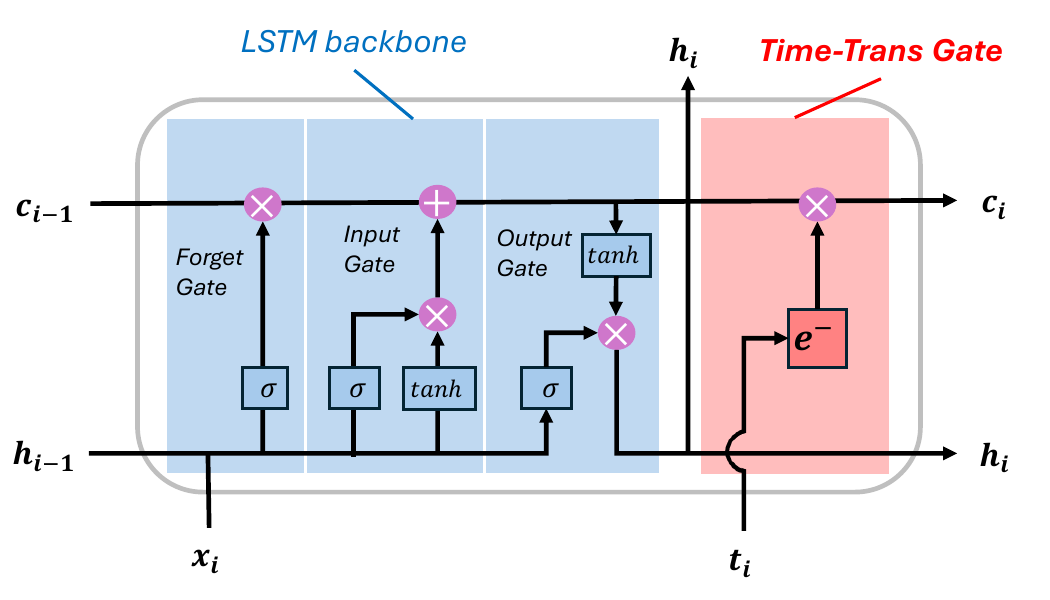}
  \caption{Unit architecture of the LITT model, where the absolute timestamp $  t  $ serves as a dedicated input to the \emph{Time-Transformation Gate}. The gate's update is separated from the standard LSTM backbone, and its output is multiplied by the cell state.}
  \vspace{-1mm}
  \label{fig:gate}
  \vspace{-1mm}
\end{figure}

\subsection{Time-Transformation Gate}

We introduce the \emph{\textbf{Time-Transformation Gate}}, built upon an LSTM recurrent backbone, as illustrated in Figure \ref{fig:gate}.
Notably, the LSTM cell state preserves complete global-dynamic information across long sequences without suffering from vanishing gradients (unlike the GRU hidden state, which is susceptible to exponential decay - see Appendix A for proof). This makes LSTM a natural choice for parameterizing time-transformation coefficients.

In contrast to applications such as language modeling or image processing, where LSTM is often regarded as merely providing longer memory than GRU without any fundamental difference \cite{yang2020lstm, zarzycki2021lstm}, in clinical time-series learning, precise timing computation is paramount. This makes LSTM the necessary architecture backbone for reliable temporal modeling, whereas GRU proves inadequate.

The LITT processing flow is then:
\begin{align}
    f_i &= \sigma(U_f h_{i-1} + W_f x_i + b_f) \quad \textit{forget gate} \nonumber \\
    I_i &= \sigma(U_I h_{i-1} + W_I x_i + b_I) \quad \textit{input gate} \nonumber \\
    o_i &= \sigma(U_o h_{i-1} + W_o x_i + b_o)  \quad \textit{output gate} \nonumber \\
    y_i &= U_h h_{i-1} + W_h x_i + b_h  \quad \textit{candidate predict} \nonumber \\
    {c}_i &= f_i \odot c_{i-1} + I_i \odot \phi(y_i) \quad \textit{ original cell state} \nonumber \\
    h_i &= o_i \odot \phi(c_i) \quad \textit{hidden state} \nonumber \\
    \gamma_i &= W_{\gamma} t_i + b_{\gamma} \quad \textit{time scaling coefficient} \nonumber\\
    c_i &= e^{-\gamma_i} c_i  \quad \textit{time-transformation coefficient } e^{-\gamma_i} \textit{for } c_i \nonumber 
\end{align}
Here, \( x_i \in \mathbb{R}^n \) represents the $n$-dimensional input feature at time step \( i = 0, \ldots, T \).
The operator \( \odot \) means element-wise multiplication. 
The nonlinear activation function \( \phi \) is typically \( \tanh(x) = (e^x - e^{-x)} / (e^x + e^{-x}) \), while the sigmoid function \( \sigma(x) = (1 + e^{-x})^{-1} \) is standard for gating nonlinearity. 

\begin{figure*}[h]
  \includegraphics[width=0.8\textwidth]{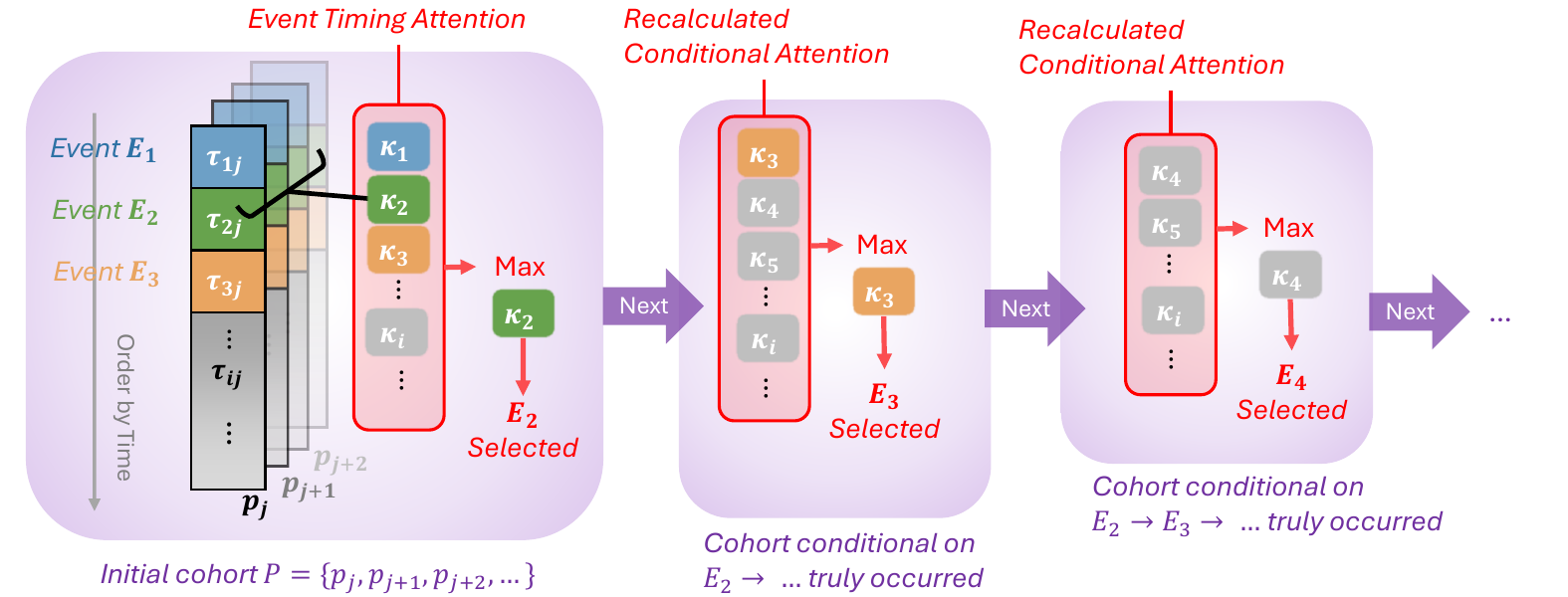}
\caption{A sequence of event-timing attention selections, resulting in the discovery of the most significant trajectory pattern $  E_2 \to E_3 \to E_4 \to \ldots  $ in patient cohort $ P= \{p_j, p_{j+1}, p_{j+2}, \ldots\}$.} \label{fig:attention}
\end{figure*}

According to their update mechanisms, the parameters are categorized into two levels:
\begin{enumerate}
\item \emph{Group-Level} parameters: $  \{W, U, b\}_{f,I,o}  $ for the three standard LSTM gates (forget, input, and output).
\item \emph{Individual-Level} parameters: matrices $  W_\gamma $ and $  b_\gamma $ that represent the time-scaling coefficients $  \gamma  $. 
\end{enumerate}

\subsection{Conditional Timing Attention}

Let $  P = \{p_j\}_{j=0}^N  $ denote the cohort of patients.
After the LITT modeling process, each patient $  p_j  $ obtains a sequence of relative timestamps $  (\tau_i)_{i=0}^T  $ corresponding to its $  T  $ potential events ordered by time. The $  i  $-th relative timestamp is computed by accumulating the preceding time-scaling coefficients. For convenience, we let $  V  $ collectively represent the parameters $  W_\gamma  $ and $  b_\gamma  $, so that:

$$\tau_i = \exp\left( -\sum_{l=0}^{i} (W_\gamma t_l + b_\gamma) \right) = \exp(-V_i T_i),$$
where $  T_i = \sum_{l=0}^{i} t_l  $ denotes the cumulative absolute time up to step $  i  $.
For any particular event, the relative timestamps across all patients exhibiting that event form a temporal distribution on the relative timeline. We quantify the central concentration of this distribution using excess kurtosis:
\vspace{-2mm}

$$\kappa(\tau_i \sim_{\text{i.i.d.}} P) = \frac{\mu_4}{\sigma^4} - 3,$$

where $  \mu_4  $ is the fourth central moment and $  \sigma  $ is the standard deviation. A higher $  \kappa  $ indicates stronger central concentration around the mean (leptokurtic behavior), accompanied by heavier tails and more extreme outliers.

Figure \ref{fig:attention} illustrates the sequential process by which the timing attention values $\kappa$ of events are selected to form a discovered trajectory. 
At each selection step, $\kappa$ values of all present events are computed across their candidate cohorts; the event with the highest $\kappa$ (indicating the greatest temporal significance) is selected as the current step  — the subsequent step is conditioned on this selection.
The candidate cohort for the next step is then restricted to patients for whom this event actually occurs as their immediate next event. 
At each step, $\kappa$ values are recomputed for all remaining events, with the distributed $\kappa$ values re-referenced to the most recently selected event as the new time-zero. The source code is available 
\footnote{\url{https://github.com/kflijia/LITT_python_code.git}}.

\section{Experiments}
To validate LITT's effectiveness in clinical practice, we use real-world longitudinal EHR time-series data from 3,276 breast cancer patients monitored between 2012 and 2024 (minimum follow-up: 1 year), originally sourced from the Fairview Health System.
The primary goal is to predict the diagnosis timing of cardiotoxicity-induced heart disease, a major complication of breast cancer treatment and one of the leading causes of mortality in this population \cite{mehta2018cardiovascular}.
Predictions target three key cardiovascular outcomes: heart failure (HF), ischemic heart disease (IS), and arrhythmias (ARR).

Input features comprise 36 dimensions extracted from structured EHR data (see Appendix B for detailed information). 
The cohort exhibits substantial temporal heterogeneity, with sequence lengths ranging from approximately 100 to over 4,000 time steps. All sequences are anchored to the breast cancer diagnosis date (day 0), with timestamps recorded as days elapsed since diagnosis.

The experiments demonstrate two featured applications of the proposed paradigm, enabled by the individual-level relative timing distribution:
\begin{enumerate}
\item Data-driven trajectory discovery
\item Counterfactual timing deduction (deducing event timing as if positive patients were negative)
\end{enumerate}
Beyond these featured applications, we evaluate LITT's group-level performance on timing prediction and single-risk survival analysis. 
The former is evaluated on the breast cancer EHR dataset, while the latter adopts two widely used public benchmarks, SUPPORT and METABRIC, to enable broader comparative evaluation against established methods.

\subsection{Data-Driven Trajectory Discovery}

\begin{figure*}[h]
  \centering
  \includegraphics[width=0.96\textwidth]{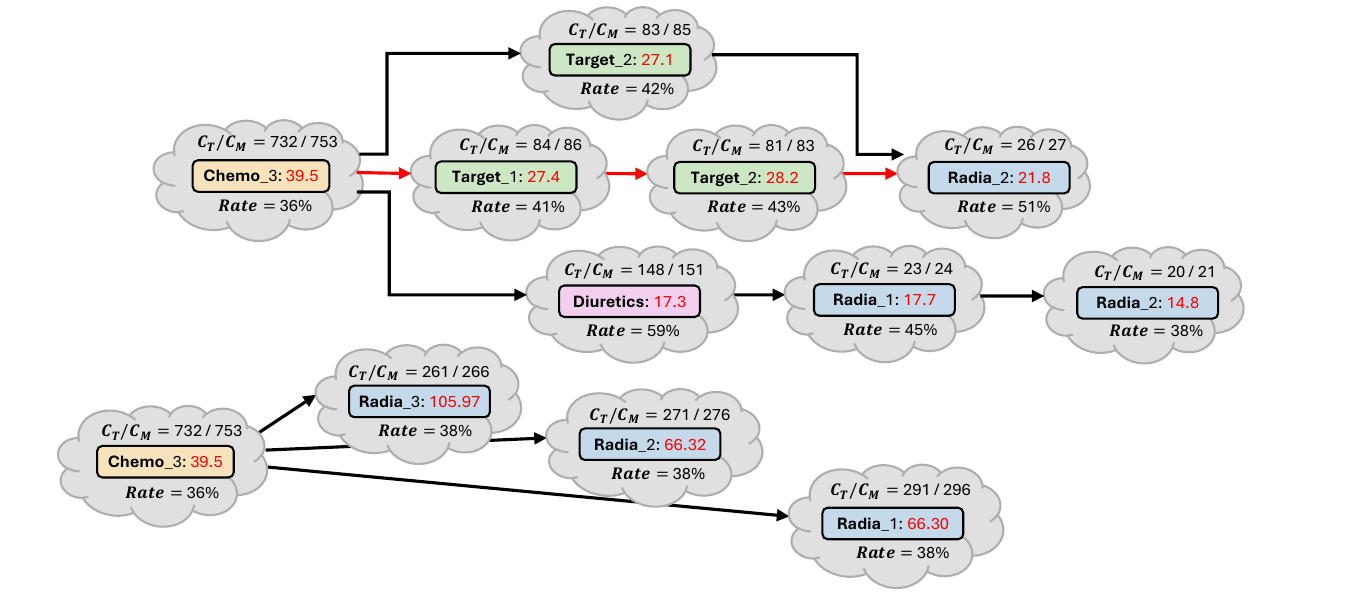}
  \vspace{-4mm}
  \caption{Two representative trajectory clusters discovered by LITT in a purely data-driven manner from real-world EHR data. Red numeric values denote the model-derived timing attention $\kappa$  of each event; $C_M$ is the count of model-derived core candidate patients (after excluding kurtosis outliers), and $C_T$ is the observed true count within $C_M$. \emph{Rate} means the positive rate of the true cohort. Numeric suffixes denote administration order (e.g., $\text{Chemo}\_3 =$ third chemotherapy). An interactive script enables full exploration across a variety of events of interest \protect\footnotemark[2].}
  \label{fig:exp1_tro}
\end{figure*}

\footnotetext[2]{\url{/github.com/kflijia/LITT_python_code/blob/main/main_LITT_Path.ipynb}}

Among the 36 features, we selected 17 binary variables to define 23 candidate events. The three major treatment procedures (Radiation, Chemotherapy, and Targeted therapy) each contribute three events that correspond to their first, second, and third administrations for the same patient. 
The remaining 14 binary features comprise 5 diagnoses of related comorbidities (e.g., hypercholesterolemia and hypertension) and 9 commonly prescribed medications for effects such as cholesterol management, blood pressure control, and cardiovascular protection.
The trajectories are detected by sequentially applying conditional timing-attention selection using an appropriate $\kappa$ threshold.

Figure \ref{fig:exp1_tro} presents two representative trajectory clusters. 
The effectiveness of LITT is reflected in three aspects:  
1) For any given event, the true cohort count $C_T$ from data remains close to the model-derived cohort count $C_M$.
2) In general, events with larger cohort sizes achieve higher $\kappa$ values,
3) Trajectories exhibit clear increasing/decreasing trends in cohort positive rates.

The top panel contains three different pathways, all originating from a common preceding event $\text{Chemo}\_3$ — the third chemotherapy administration.
The first two pathways share identical terminal events ($\text{Radia}\_2$) and exhibit closely matched values of attention $\kappa$, cohort count, and positive rate; they are therefore merged into a single representation. 
Their convergence at a shared endpoint suggests that the intermediate segments of these two pathways carry equivalent temporal significance. Since the first pathway follows ``$... \rightarrow \text{Target}\_2 \rightarrow ... $'' and the second follows ``$... \rightarrow \text{Target}\_1 \rightarrow \text{Target}\_2 \rightarrow ... $'', this indicates that the first administration of the targeted therapy, $\text{Target}\_1$, is temporally redundant, since its significance is subsumed by the second administration, $\text{Target}\_2$. 

Interestingly, the third pathway also terminates at $\text{Radia}\_2$ but cannot be merged with the other two. Within this pathway, the event $\text{Diuretics}$ — the first administration of diuretics, a medication used to support kidney function and manage heart failure — acts as a critical turning point. 
Accordingly, the cohort positive rate peaks at this step ($59\%$) and declines progressively at subsequent steps, a trend that stands in clear contrast to the monotonically increasing pattern observed in the other two pathways.


The bottom panel contains three two-step pathways, all exhibiting significantly high $\kappa$ values. 
All three originate from $\text{Chemo}\_3$ but terminate at the third, second, and first radiation administrations ($\text{Radia}\_3$, $\text{Radia}\_2$, and $\text{Radia}\_1$), respectively, with $\kappa$ values declining in the same order.
Notably, the proposed timing-attention mechanism enables concurrent evaluation of events without imposing any logical ordering constraint — a substantial efficiency advantage over traditional association rule-based approaches, which incur exponential search complexity. 
In this case, $\text{Radia}\_3$ achieves a markedly higher $\kappa$ than $\text{Radia}\_2$ and $\text{Radia}\_1$, suggesting greater temporal significance. However, since all three events share an equal positive rate $38\%$, this does not imply that a greater number of radiation administrations is more predictive of heart disease; rather, it reflects the fact that most patients in this cohort ultimately undergo three rounds of radiation treatment.



\begin{figure*}[h]
  \centering
  \includegraphics[width=0.99\textwidth]{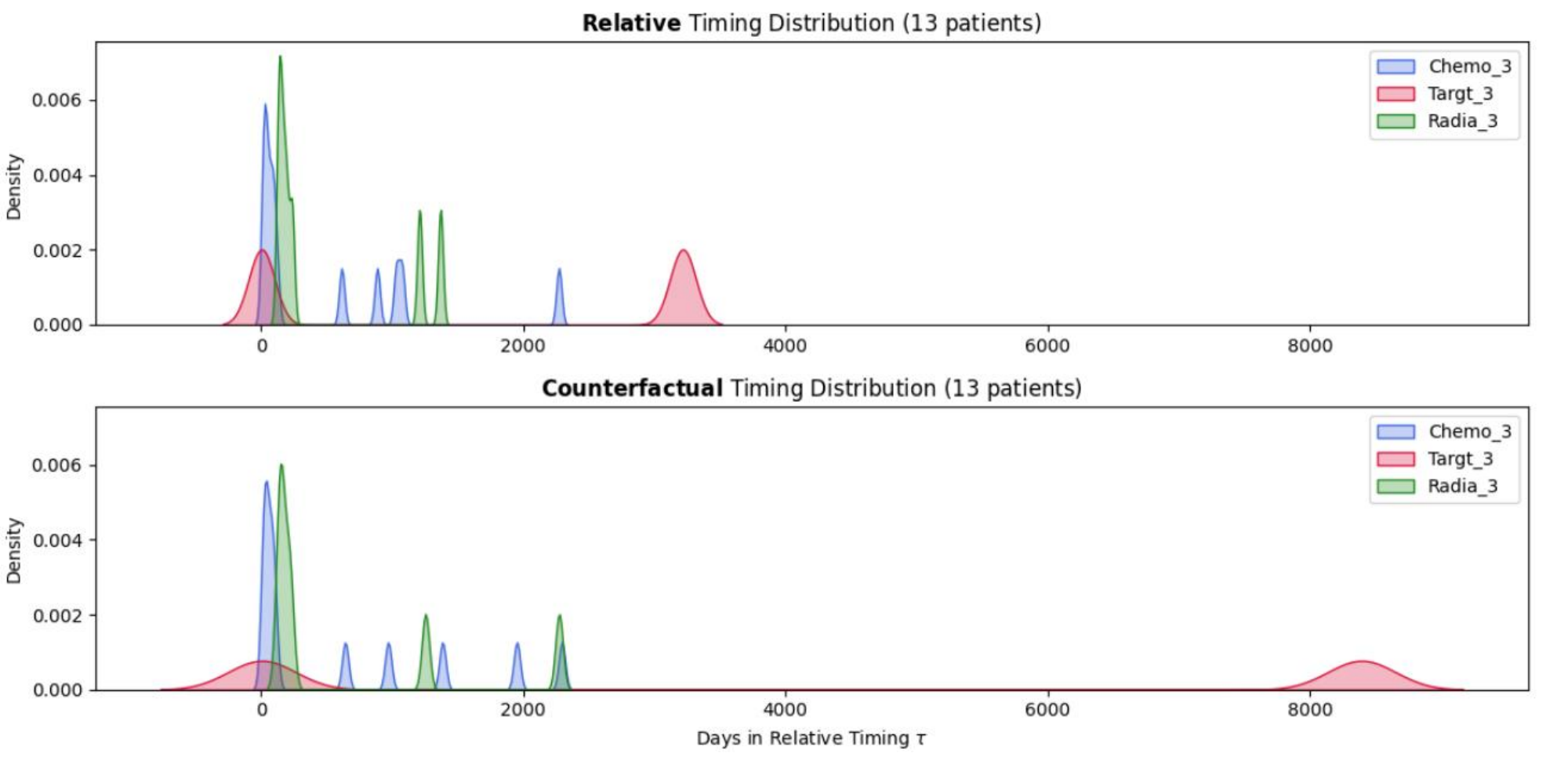}
  \vspace{-4mm}
  \caption{KDE plots of model-derived relative and counterfactual timing distributions for three events ($\text{Chemo}\_3$, $\text{Target}\_3$, and $\text{Radia}\_3$), under the counterfactual assumption that positive patients following the trajectory $\text{Dx\_DB}\_2 \rightarrow \text{Dx\_HT}\_3 \rightarrow \text{Chemo}\_3$ remain negative with respect to cardiotoxicity. $\text{Dx\_DB}$ and $\text{Dx\_HT}$ denote diagnoses of diabetes and hypertension, respectively. An interactive script enables further exploration of events along trajectories of interest \protect\footnotemark. }
  \label{fig:exp2_flip}
\end{figure*}

\footnotetext{\url{/github.com/kflijia/LITT_python_code/blob/main/main_LITT_Flipped.ipynb}}

\subsection{Counterfactual Timing Deduction }
In this experiment, we recalculate the relative timing distribution of a selected cohort by reassigning their observed labels from positive to negative. 
The group-level parameters remain fixed throughout; only the individual-level temporal variable $\gamma$ is retrained, ensuring that the derived counterfactual timing distribution remains \emph{comparable} to the original within the same relative timeline $\tau$. 
This controlled reassignment enables data-driven causal deduction by isolating the effect of label status at the individual level.

Figure \ref{fig:exp2_flip} presents a representative comparison of the original and counterfactual timing distributions within the shared relative timeline $\tau$. We observe that for approximately half of the cohort, the timing of $\text{Target}\_3$ shifts markedly toward larger values under the counterfactual assumption, suggesting that postponing (or canceling if the counterfactual timing exceeds the patient's observation window in $\tau$) the third targeted therapy could substantially reduce the risk of cardiotoxicity onset for these patients.


\subsection{Timing Prediction \& Survival Analysis }

\begin{figure*}[h]
  \centering
  \includegraphics[width=0.98\textwidth]{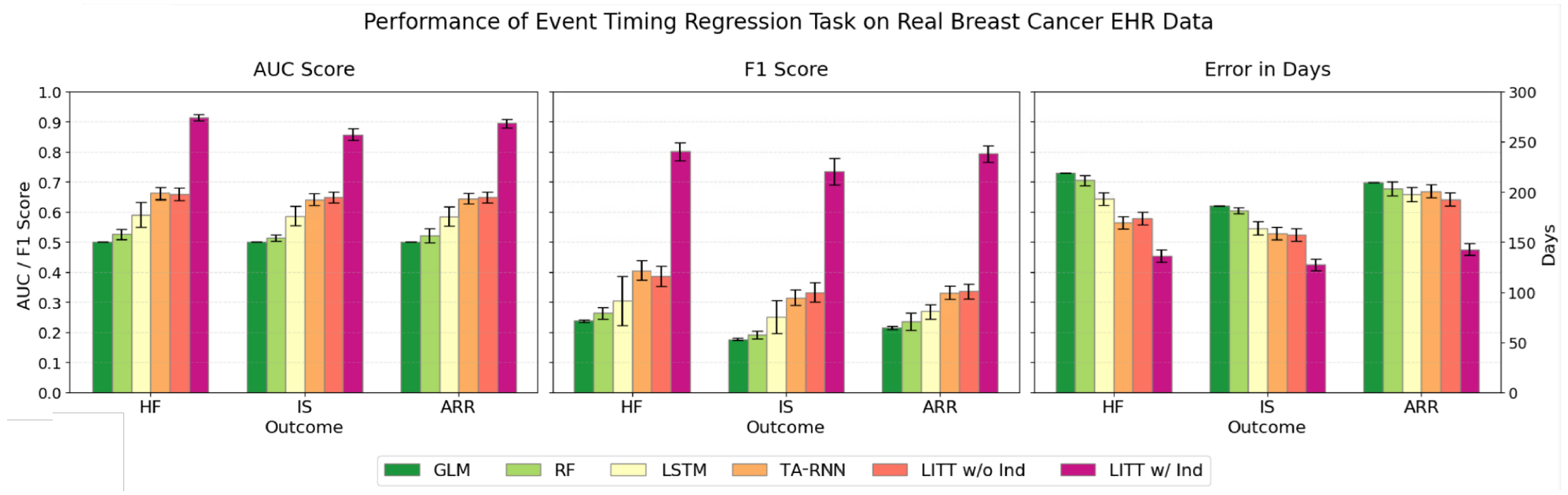}
  \vspace{-2mm}
\caption{Performance comparison across three cardiovascular outcomes: heart failure (HF), ischemic heart disease (IS), and arrhythmias (ARR).
LITT w/o Ind denotes the ablated variant in which all individual-level time-scaling coefficients $\gamma$ are zeroed out, reducing the model to a group-level baseline. LITT's averaged group-level performance is comparable to TA-RNN and superior to standard LSTM, confirming that the group-level component of LITT is competitive independently of the individual-level temporal features.}
  \label{fig:exp2_bar}
\end{figure*}




For a more intuitive evaluation, Figure \ref{fig:exp2_bar} also reports the AUC and F1 score alongside the RMSE in number of days. These binary metrics are derived by thresholding predicted event days against each patient's last observed day — predictions beyond the observation window are treated as negative, earlier predictions as positive. Note that since these metrics are derived from a regression output rather than direct binary classification, they are not directly comparable to dedicated classification models.

In addition to the standard LSTM and AT-RNN \cite{al2024ta}, we also include linear logistic regression (GLM) and random forest (RF) as complementary baselines for evaluation.
The LITT method is evaluated in two configurations: with and without activation of the time-scaling coefficients $\gamma$, which capture individual-level temporal features.
The substantial performance gap between these two modes indicates that most of the individualized temporal information is derived from the ground-truth outcome in days rather than longitudinal data. Consequently, achieving optimal predictive performance with LITT may require additional external patient-specific data.

\begin{table}[h]
\centering
\small
\begin{tabular}{|
>{\columncolor[HTML]{DCDCDC}}l |l|l|}
\hline
\textbf{Methods}      & \cellcolor[HTML]{DCDCDC}\textit{SUPPORT} & \cellcolor[HTML]{DCDCDC}\textit{METABRIC} \\ \hline
\textbf{CPH}          & 0.596 ($\pm$ 0.004)                            & 0.623 ($\pm$0.020)                             \\ \hline
\textbf{RSF}          & 0.639 ($\pm$0.005)                            & 0.665 ($\pm$0.022)                             \\ \hline
\textbf{DeepSurv}     & 0.619 ($\pm$0.005)                            & 0.641 ($\pm$0.022)                             \\ \hline
\textbf{DeepHit}      & 0.630 ($\pm$0.007)                            & 0.642 ($\pm$0.023)                             \\ \hline
\textbf{DSM}          & 0.637 ($\pm$0.005)                            & 0.670 ($\pm$0.014)                             \\ \hline
\textbf{SurvTrace}    & 0.649 ($\pm$0.004)                            & \textbf{0.682 ($\pm$0.115)}                    \\ \hline
\textbf{LITT w/o Ind} & \textbf{0.669 ($\pm$0.006)}                   & 0.673 ($\pm$0.023)                             \\ \hline
\end{tabular}
\vspace{3mm}
\caption{C-index performance of single-risk survival analysis on public datasets SUPPORT and METABRIC.} 
\label{tab}
\vspace{-6mm}
\end{table}


For survival analysis, the C-index is derived from LITT's predicted event timings for evaluation. We compare LITT against several state-of-the-art deep survival analysis methods: DeepHit \cite{lee2018deephit}, DeepSurv \cite{katzman2018deepsurv}, SurvTrace \cite{wang2022survtrace}, and Deep Survival Machine (DSM) \cite{nagpal2021deep}. Additionally, the Cox proportional hazards (CPH) model and random survival forest (RSF) are included as classical baselines. 

For comparability, all models are evaluated at the 50\% survival probability threshold. As shown in Table \ref{tab}, the ablated LITT variant — with individual-level time-scaling coefficients $\gamma$ zeroed out — achieves competitive results across methods, though performance varies by dataset and is not consistently the highest.

\section{Conclusion}

This work introduces LITT, a data-driven paradigm for individual-level temporal trajectory modeling. By formally defining Timing Attention, we establish the conceptual prerequisites for timing computation absent from the existing literature, and empirically demonstrate automatic trajectory discovery and counterfactual timing deduction (i.e., a What-If Machine) without recourse to prior domain knowledge. We further establish that LSTM's dedicated cell state is architecturally essential for timing computation, a distinction from GRU that has been largely overlooked.

Several directions remain open. The event-first perspective of discovering events post-hoc from collective timing distributions has not yet been fully realized and represents a natural next step. The individual-level $\tau - t$ residual warrants further investigation as a personalized prognostic feature, and aggregating across multiple training runs opens the door to possibility-space-grounded survival analysis conclusions.
At a deeper level, this work points toward a reformulation of causal theory, in which causality is defined over an open possibility space rather than a closed set of variables.

This work paves the way for more temporally faithful deep learning models in digital healthcare, establishing foundational building blocks for future progress in precision medicine and causal-reasoning AI.

\bibliographystyle{ACM-Reference-Format}
\bibliography{main}
\appendix

\section{Appendix A: Why LSTM Enables Timing Computation While GRU Cannot}
\vspace{2mm}

Two prominent recurrent architectures, LSTM (Long Short-Term Memory) and GRU (Gated Recurrent Unit), have been extensively compared on tasks such as language modeling and computer vision, which usually have limited temporal dependencies, with a focus on computational efficiency \ cite {chung2014empirical, zarzycki2021lstm}.

However, in the context of time-series learning, LSTM has been severely underestimated; its dedicated cell state, which preserves temporally consistent status, is essential for timing computation across the global timeline.

On the contrary, GRUs lack such a capability.
The distinction lies in their internal state update mechanisms.

LSTM's processing flow and state update Jacobian are:
\begin{align*}
    f_i &= \sigma(U_f h_{i-1} + W_f x_i + b_f) \quad \textit{forget gate} \quad\\
    I_i &= \sigma(U_i h_{i-1} + W_i x_i + b_i) \quad \textit{input gate}\\
    o_i &= \sigma(U_o h_{i-1} + W_o x_i + b_o)  \quad \textit{output gate}\\
    y_i &= U_h h_{i-1} + W_h x_i + b_h  \quad \textit{candidate}\\
    {c}_i &= f_i \odot c_{i-1} + i_i \odot \phi(y_i) \quad \textit{ \textbf{cell state}}\\
    h_i &= o_i \odot \phi(c_i) \quad \textit{\textbf{hidden state}} \\
    J_c &= \frac{\partial c_i}{\partial c_{i-1}} = \text{diag}(f_i) \quad \\
    J_h &= \text{diag}(o_i) \cdot \text{diag}(1 - \phi^2(c_i)) \cdot \frac{\partial c_i}{\partial h_{i-1}}
    \\
\end{align*}

GRU's processing flow and state update Jacobian are:
\begin{align*}
    z_i &= \sigma(W_z x_i + U_z h_{i-1} + b_z) \quad \textit{update gate}\\
  r_i &= \sigma(W_r x_i + U_r h_{i-1} + b_r) \quad \textit{reset gate}\\
  \tilde{h}_i &= \phi(W_h x_i + U_h (r_i \odot h_{i-1}) + b_h) \textit{ candidate } \\
  h_i &= (1 - z_i) \odot h_{i-1} + z_i \odot \tilde{h}_i   \quad \textit{\textbf{hidden state}}\\
J_h &=\frac{\partial h_i}{\partial h_{i-1}} \\
&= \text{diag}(1 - z_i) + \text{diag}(z_i) \cdot \text{diag}(1 - \tilde{h}_i^2) \cdot U_h \cdot \text{diag}(r_i) \\
\end{align*}

\vspace{-2mm}
GRUs maintain a single hidden state only, updated as a convex combination of the previous hidden state and a candidate activation, which leads to multiplicative forgetting as the blend approaches the candidate.

In contrast, LSTM's separate cell state uses additive updates, with the forget gate $f_i\in[0,1]$ controlling retention, ensuring stable gradients over long sequences.
Analysis of their state-update Jacobians reveals that, in both architectures, the hidden state is susceptible to exponential decay due to scaling terms like \( 1 - \tilde{h}_i^2 \in (0, 1] \), whereas the LSTM cell state $c_t$ supports additive accumulation without this limitation.

LSTM's additive cell-state updates allow all historical states to contribute equally when assessing mutual dependencies, enabling the alignment of identical events across varying timings.
While LSTMs excel at capturing knowledge, GRUs prioritize simulating empirical patterns that mimic human memory forgetting.

\section{Appendix B: Experimental Data Features}

\begin{table}[H]
\centering
\small 
\caption{Description of outcome features}
\begin{tblr}{
  row{1} = {c},
  cell{2}{1} = {r=3}{},
  cell{5}{1} = {r=4}{},
  vlines,
  hline{1,10} = {-}{0.08em},
  hline{2,5,9} = {-}{},
  hline{3-4,6-8} = {2-3}{},
}
\textbf{\makecell{Outcome }} & \textbf{\makecell{Included diseases}} & \textbf{ICD code}\\
\makecell{Heart \\ Failure (HF)} & \makecell{ Combine
Heart \\ Failure (CHF)} & \makecell{ICD9
= 428.*; ICD10 = I50.*}\\
 & Cardiomyopathy
(CM) & \makecell{ICD9 = 425.*; ICD10 = I42.*}\\
 & Pulmonary
Edema & \makecell{ICD9 = 514.*; ICD10 = J81.*}\\
\makecell{Ischemic \\
Heart \\ Disease (IS)}~ & \makecell{Myocardial \\
Infarction (MI)} & {\makecell{ICD9
= 410.*, 411.*, 412.*~~~~~~\\~ICD10 = I21.*,
I22.*, I23.*}}\\
 & Angina
Pectoris & \makecell{ICD9
= 413.*; ICD10 = I20.*}\\
 & \makecell{Coronary
Artery \\ Disease (CAD)} & \makecell{ICD9
= 414.*\\ ICD10 = I24.*}\\
 & \makecell{Pulmonary \\
Heart Disease} & \makecell{ICD9
= 415.*\\ ICD10 = I25.*}\\
\makecell{Arrhythmias \\ (ARR)} &  & {ICD9 = 427.9*~\\~ICD10 = I48.*, I49.*~ ~ ~ ~}
\end{tblr}
\label{tab:outcomes}
\end{table}

The detailed interpretation of the outcome variables is provided in Table \ref{tab:outcomes}. Among the 36 clinical features selected as covariates, demographic and vital-sign features include: age, race, BMI, weight, height, SBP (Systolic Blood Pressure), DBP (Diastolic Blood Pressure), and heart rate. Additional features are detailed in Table \ref{tab:vars}.

\onecolumn

\begin{longtblr}[
  caption = {Description of modeling features\label{tab:vars}}
]{
  row{1} = {c},
  hlines,
  vlines,
  vline{-} = {1}{0.08em},
  hline{1-2,30} = {-}{0.08em},
}
\small
\textbf{Variable} & \textbf{Type} & \textbf{Description}\\
LDL(Low-Density Lipoprotein) & numeric~ ~ ~ & LOINC test codes 2089-1 and 13457-7\\
HDL(High-Density Lipoprotein) & numeric~ ~ ~ & LOINC test codes 2085-9 and 43396-1\\
HBA1C(hemoglobin A1C Test) & numeric~ ~~ & LOINC test code equal to 4548-4\\
Troponin & numeric~ ~ & LOINC test codes 10839-9, 42757-5, and 89579-7~ ~ ~\\
BNP(brain natriuretic peptide) & numeric~ ~ & LOINC test codes 30934-4 and 33762-6~ ~ ~\\
WBC(White Blood Cell) & numeric~ ~ & {LOINC test codes~6690-2,~26464-8,~729-4,~49498-9, \\and 5820-6~}\\
Abnormal Glucose & numeric~ ~ & The Glucose value with abnormal flag = 1\\
Abnormal Creatinine & numeric~ ~ & The Creatinine value with abnormal flag = 1\\
Anthracyclines & binary & {whether having anthracyclines-based chemotherapy \\medicine using}\\
Targeted~ ~ ~ & binary & whether having targeted therapy medicine using\\
Radiation~ ~ ~ & binary & {whether having radiation treatment, including \\procedure names:~~Radiation, IMRT,~Proton,~SRS,~\\Hyperthermia,~IORT, and~Intracavitary radiation ~~}\\
Insulin (Md\_Insl)~ ~ ~ & numeric & Diabetes medicine\\
Metformin (Md\_Metf) & numeric & T2DM~medicine\\
Sulfonylureas (Md\_Sulf) & numeric & T2DM old drug\\
SGLT2inhibitor  (Md\_SGLT2) & binary & Medicine for T2DM, heart failure, CKD\\
GLP-1Receptor Agonists & binary & {Medicine for T2DM, weight loss, cardiovascular protection}\\
Ezetimibe  (Md\_Ezt) & binary & Cholesterol-lowering medication\\
PCSK9inhibitor & binary & Newer cholesterol-lowering biologic drugs\\
Fibrates  (Md\_Fibr) & binary & Lipid-lowering medications\\
BileAcid Sequestrants  (Md\_BAS) & binary & Lipid-lowering medications\\
Niacin & binary & Lipid-lowering agent\\
Diuretics  (Md\_Diu)~ ~ ~ & binary & {Medicine for~hypertension, heart failure, edema, and \\sometimes kidney or liver disorders ~}\\
Hyperlipidemia & binary & {Whether having Hyperlipidemia diagnosed:\\ICD9 codes 272.2, 272.4, 272.5\\ICD10 codes E78.2, E78.4, E78.5}\\
Diabetes (Dx\_DB) & binary & {Whether having Diabetes diagnosed:\\ICD9 codes 250.* or ICD10 codes E11.*}\\
Hypertension (Dx\_HT) & binary & {Whether having Hypertension diagnosed:\\ICD9 codes in the range from 401.* to 405.*\\ICD10 codes in the range from I10.*to I16.*}\\
Smoking (Dx\_Smk) & binary & {ICD-9 codes 305.1* and V15.82\\ICD-10 code Z87.*}\\
Obesity (Dx\_Ob) & binary & ICD-9 codes 278.*~ ICD-10 code E66.*\\
Chronic Kidney Disease (Dx\_CKD) & binary & ICD-9 codes 585.*~ ICD-10 code N17.*
\end{longtblr}

\section{Appendix C: Numerical Results in Figure 6}

\begin{table}[H]
\small
\begin{tabular}{|l|l|l|l|}
\hline
\textbf{AUC}                         & \cellcolor[HTML]{D7D6D6}HF & \cellcolor[HTML]{D7D6D6}IS & \cellcolor[HTML]{D7D6D6}ARR \\ \hline
\cellcolor[HTML]{D7D6D6}GLM          & 0.5 $\pm$ 0                & 0.5 $\pm$ 0                & 0.5 $\pm$ 0                 \\ \hline
\cellcolor[HTML]{D7D6D6}RF           & 0.525 $\pm$ 0.016          & 0.514 $\pm$ 0.010          & 0.522 $\pm$ 0.244           \\ \hline
\cellcolor[HTML]{D7D6D6}LSTM         & 0.592 $\pm$ 0.042          & 0.587 $\pm$ 0.032          & 0.586 $\pm$ 0.030           \\ \hline
\cellcolor[HTML]{D7D6D6}TA-RNN          & 0.661 $\pm$ 0.022          & 0.638 $\pm$ 0.020          & 0.640 $\pm$ 0.016           \\ \hline
\cellcolor[HTML]{D7D6D6}LITT w/o Ind & 0.659 $\pm$ 0.021          & 0.650 $\pm$ 0.018          & 0.648 $\pm$ 0.018           \\ \hline
\cellcolor[HTML]{D7D6D6}LITT w/ Ind  & 0.914 $\pm$ 0.011          & 0.858 $\pm$ 0.019          & 0.895 $\pm$ 0.014           \\ \hline
\end{tabular}
\end{table}

\vspace{-3mm}
\begin{table}[H]
\begin{tabular}{|l|l|l|l|}
\hline
\textbf{F1}                          & \cellcolor[HTML]{D7D6D6}HF & \cellcolor[HTML]{D7D6D6}IS & \cellcolor[HTML]{D7D6D6}ARR \\ \hline
\cellcolor[HTML]{D7D6D6}GLM          & 0.238 $\pm$ 0.004          & 0.177 $\pm$ 0.004          & 0.214 $\pm$ 0.005           \\ \hline
\cellcolor[HTML]{D7D6D6}RF           & 0.264 $\pm$ 0.020          & 0.191 $\pm$ 0.014          & 0.235 $\pm$ 0.029           \\ \hline
\cellcolor[HTML]{D7D6D6}LSTM         & 0.304 $\pm$ 0.081          & 0.251 $\pm$ 0.055          & 0.269 $\pm$ 0.023           \\ \hline
\cellcolor[HTML]{D7D6D6}TA-RNN          & 0.402 $\pm$ 0.031          & 0.312 $\pm$ 0.028          & 0.331 $\pm$ 0.022           \\ \hline
\cellcolor[HTML]{D7D6D6}LITT w/o Ind & 0.387 $\pm$ 0.034          & 0.333 $\pm$ 0.032          & 0.336$\pm$ 0.024            \\ \hline
\cellcolor[HTML]{D7D6D6}LITT w/ Ind  & 0.801 $\pm$ 0.029          & 0.735 $\pm$ 0.043          & 0.794 $\pm$ 0.027           \\ \hline
\end{tabular}
\end{table}

\vspace{-3mm}
\begin{table}[H]
\begin{tabular}{|l|l|l|l|}
\hline
\textbf{Days}                        & \cellcolor[HTML]{D7D6D6}HF & \cellcolor[HTML]{D7D6D6}IS & \cellcolor[HTML]{D7D6D6}ARR \\ \hline
\cellcolor[HTML]{D7D6D6}GLM          & 219 $\pm$ 0                & 185 $\pm$ 0                & 209 $\pm$ 0                 \\ \hline
\cellcolor[HTML]{D7D6D6}RF           & 211 $\pm$ 4                & 181 $\pm$ 3                & 203 $\pm$ 7                 \\ \hline
\cellcolor[HTML]{D7D6D6}LSTM         & 193 $\pm$ 6                & 163 $\pm$ 6                & 198 $\pm$ 6                 \\ \hline
\cellcolor[HTML]{D7D6D6}TA-RNN          & 169 $\pm$ 6                & 158 $\pm$ 6                & 201 $\pm$ 6                 \\ \hline
\cellcolor[HTML]{D7D6D6}LITT w/o Ind & 173 $\pm$ 6                & 157 $\pm$ 6                & 192 $\pm$ 6                 \\ \hline
\cellcolor[HTML]{D7D6D6}LITT w/ Ind  & 136 $\pm$ 5                & 127 $\pm$ 5                & 142 $\pm$ 6                 \\ \hline
\end{tabular}
\end{table}

\end{document}